\documentclass{article}

\usepackage[preprint, nonatbib]{neurips_2019}

\usepackage[utf8]{inputenc} 
\usepackage[T1]{fontenc}    
\usepackage{hyperref}       
\usepackage{url}            
\usepackage{booktabs}       
\usepackage{amsfonts}       
\usepackage{nicefrac}       
\usepackage{microtype}      
\include{macros_new}
\include{new_cmds}
\usepackage{epstopdf}
\usepackage{tikz} 
\usepackage{pgfplots}
\pgfplotsset{compat=newest} 
\usepgfplotslibrary{units} 
\usepackage{subcaption}
\usepackage{siunitx}
\usepackage{wrapfig}
\usepackage{authblk}
\usepackage[english]{babel}
\usepackage{blindtext}
\title{A Study of BFLOAT16 for Deep Learning Training}

\author[1]{\textbf{Dhiraj Kalamkar}}
\author[2]{\textbf{Dheevatsa Mudigere}}
\author[1]{\textbf{Naveen Mellempudi}\thanks{corresponding author \texttt{\{naveen.k.mellempudi\}@intel.com}}\hspace{4pt}}
\author[1]{\textbf{Dipankar Das}}
\author[1]{\\ \textbf{Kunal Banerjee}} 
\author[1]{\textbf{Sasikanth Avancha}}
\author[1]{\textbf{Dharma Teja Vooturi }\thanks{IIIT Hyderabad, $^\ddagger$Lab126 Amazon, work done while at at Intel}\hspace{4pt}}
\author[1]{\textbf{Nataraj Jammalamadaka}$^\ddagger$}
\author[2]{\textbf{Jianyu Huang}}
\author[2]{\textbf{Hector Yuen}}
\author[2]{\textbf{Jiyan Yang}}
\author[2]{\textbf{Jongsoo Park}}
\author[1]{\textbf{Alexander Heinecke}}
\author[1]{\\ \textbf{Evangelos Georganas}}
\author[1]{\textbf{Sudarshan Srinivasan}}
\author[1]{\textbf{Abhisek Kundu}}
\author[2]{\\ \textbf{Misha Smelyanskiy}}
\author[1]{\textbf{Bharat Kaul}}
\author[1]{\textbf{Pradeep Dubey}}
\affil[1]{Parallel Computing Lab, Intel Labs}
\affil[2]{Facebook, 1 Hacker Way, Menlo Park, CA}

\begin{document}
\vspace{-15pt}
\maketitle
\begin{abstract}
This paper presents the first comprehensive empirical study demonstrating the efficacy of the Brain Floating Point (BFLOAT16) half-precision format for Deep Learning training across image classification, speech recognition, language modeling, generative networks and industrial recommendation systems. BFLOAT16 is attractive for Deep Learning training for two reasons: the range of values it can represent is the same as that of IEEE 754 floating-point format (FP32) and conversion to/from FP32 is simple. Maintaining the same range as FP32 is important to ensure that no hyper-parameter tuning is required for convergence; e.g., IEEE 754 compliant half-precision floating point (FP16) requires hyper-parameter tuning. In this paper, we discuss the flow of tensors and various key operations in mixed precision training, and delve into details of operations, such as the rounding modes for converting FP32 tensors to BFLOAT16. We have implemented a method to emulate BFLOAT16 operations in Tensorflow, Caffe2, IntelCaffe, and Neon for our experiments. Our results show that deep learning training using BFLOAT16 tensors achieves the same state-of-the-art (SOTA) results across domains as FP32 tensors in the same number of iterations and with no changes to hyper-parameters. 

\end{abstract}

\section{Introduction} 
\label{intro}

The spectacular success of Deep Learning has come riding on the ready availability of data and a tremendous growth in compute capability of deep learning systems.
In recent years, compute growth has been driven by specialized architectures for GEMM (General Matrix Multiply) acceleration, and the shift to low precision compute.
Inference has witnessed a proliferation of mixed precision compute \cite{deepcompression,rastegariECCV16,mellempudi2017ternary,trn} where different operations execute at different precision, all the way from binary/ternary operands to 16b floating point.
Similarly, training has also witnessed its share of mixed precision methods, where a combination of half- and single-precision compute is used.

There are at least three half-precision formats in the domain of mixed precision training of large neural networks: FP16 \cite{nvbaidu_mixed}, 16-bit Integer based \cite{mixedPrec18}, and BFLOAT16 \cite{distbelief, tf_whitepaper, cloudTPU}. All these methods have 16-bit input operands and 32-bit accumulators for all the computations. Of the three formats, only the first two have publicly available description of training methodology and experimental results on a wide variety of neural networks although BFLOAT16 was originally conceived for deep learning training. BFLOAT16 data format was first introduced as part of distributed training frameworks DistBelief~\cite{distbelief} and Tensorflow~\cite{tf_whitepaper} as a low precision storage format used to reduce communication volumes of weights and activations shared between compute nodes during distributed parallel training\footnote{special thanks to Jeff Dean for pointing this out}. It has since become an alternative numeric format specifically targeted towards accelerating deep leaning training (mainly within the Google ecosystem~\cite{cloudTPU}), because of its wider dynamic range and smaller footprint. 
In this work, we present a detailed mixed precision methodology using BFLOAT16 and demonstrate coverage by training a variety of workloads from image processing (including GANs), to speech/language processing, and recommendation systems. For our experiments we employ a method, where FP32 operations emulate the behavior of BFLOAT16 operations by appropriately zeroing out the lower 16 bits and appropriately rounding the input operands. 

We have developed a library called Quantlib to implement the emulation in multiple deep learning frameworks such as IntelCaffe, Caffe2, Neon and Tensorflow. One of the functions Quantlib  provides is appropriately modifying the elements of an input FP32 tensor to emulate the behavior of BFLOAT16. Specifically, it zeroes out the lower 16 bits of the FP32 elements and performs RNE (Round to Nearest Even) rounding based on those bits. This modification ensures that the tensor possesses FP32 precision (so that FP32-hardware and FP32-libraries can operate on it), and also provides the exact precision and rounding as would be afforded by BFLOAT16 hardware.  
Quantlib is called prior to GEMM operations (or other operations which are planned to be implemented in BFLOAT16) to emulate the behavior of BFLOAT16 input operands, while FP32 output of the GEMM naturally fits into the ``BFLOAT16-input, FP32-accumulator'' schema of this training methodology.   

Using multiple deep learning frameworks modified to insert appropriate Quantlib calls, we provide SOTA results for: AlexNet \cite{alexnet}, ResNet-50 \cite{resnet}, DC-GAN \cite{dcgan}, SR-GAN \cite{ledig2017photo}, DeepSpeech2 \cite{amodei2016deep}, GNMT \cite{wu2016google}, and two industrial workloads namely: a Deep and Cross Network, and a DNN Recommendation System.
We also qualitatively compare and contrast the training methodologies using BFLOAT16, FP16 and INT16. We observe that FP16-based training requires tuning an additional hyper-parameter for loss scaling to achieve SOTA results; INT16-based training requires fine grained block-quantization and maintaining block-level scaling factors to achieve SOTA results.
In comparison all BFLOAT16 experiments are performed without any hyperparameter changes and BFLOAT16 kernels are expected to be relatively straightforward.




The rest of the paper is organized as follows.
Section \ref{relatedwork} provides a survey of the literature and describes various attempts at half-precision based training. 
Section~\ref{numerics} discusses the BFLOAT16 format, operations and data flow in detail. Section~\ref{results} describes our experimental results in detail. Section~\ref{conclusion} discusses our concluding thoughts.

\section{Related Work}\label{relatedwork}
Application of low precision datatype in deep learning is a well explored topic in research. Literature shows that various different reduced precision data representations have been investigated which can be broadly classified into two types: the more standard floating-point based formats \cite{nvbaidu_mixed,borisgtc17fp16,dettmers20158} and custom fixed point based formats \cite{vanhoucke2011improving,courbariaux2014training,gupta2015lp,hubara2016qnn,flexpoint}. 

Custom fixed point representations may offer more flexibility than typical floating point based ones in terms of both increased precision and dynamic range by maintaining separate integer values for precision and range. Consequently, fixed point representations may provide more robust and accurate training of an underlying application. The work reported in \cite{vanhoucke2011improving} leverages the dynamically scaled fixed point representation proposed in \cite{williamson1991dynamically} to speed up convolution neural networks by $4\times$ over an optimized floating point implementation on general purpose CPU hardware. In \cite{gupta2015lp}, the authors present a comprehensive study on the effect of low precision fixed point computation for deep learning. They also train smaller networks using 16-bit fixed point on specialized hardware.

Researchers have ventured into less than 16-bit precision as well and almost all of them use custom fixed point schemes. Reference \cite{courbariaux2014training} uses a dynamical fixed point format with low precision multiplications with up to 12-bit operations. This idea is further advanced in \cite{courbariaux2015binaryconnect} where the authors showcase training with only binary weights while keeping all other tensors and operations in full precision. Another extension \cite{hubara2016binarized} uses binary activations as well; however, the gradients and the weights are maintained in full precision. Another related work \cite{hubara2016qnn} uses activations and weights quantized into 6-bits for neural network training with gradients in full precision. The method described in \cite{rastegariECCV16} uses binary representation for all components including gradients. However, all the aforementioned methods are shown to work for smaller benchmark model/data-sets only and invariably result in a non-trivial drop in accuracy with larger ImageNet data-set \cite{imagenet_data} and classification task \cite{imagenet_contest}. The authors of \cite{flexpoint} advocate for a fixed point numerical format called Flexpoint that is specifically tailored for executing deep neural networks on a specialized hardware; this datatype is shown to outperform FP16 and achieve numerical parity with FP32 across a diverse set of workloads. A more general dynamic fixed point representation and associated compute primitives are presented in \cite{mixedPrec18}, which leverages general purpose hardware using the integer-compute pipeline to match FP32 baseline accuracy across state of the art convolution neural networks. 

The disadvantage of these integer based representations in contrast to floating point is the additional overheads of handling shared exponents and managing accumulator overflow. Köster et al.\cite{flexpoint} proposed an algorithm that predicts the shared exponent ahead of time to eliminate some of these overheads. However this solution requires collection of additional statistics at each layer, which cannot be efficiently computed on general purpose hardware. 


A mixed precision training methodology using FP16 and FP32 is reported in \cite{nvbaidu_mixed}. This work employs FP16 for storing activations, weights and gradients. The computations during forward pass and back propagation use FP16 datatype while results are accumulated into FP32. A master copy of the FP32 weights are preserved for the update operation. The authors successfully perform deep learning training on a wide range of applications encompassing deep networks and larger data-sets (ILSVRC-class problems) at the expense of minimal loss compared to baseline FP32 results. This work, however, underlines that FP16/FP32 mixed precision training entails loss scaling \cite{borisgtc17fp16} to attain near-SOTA results. Loss scaling, basically, ensures that back-propagated gradient values are shifted into a range which can be represented by FP16 and therefore the small magnitude (negative exponent) values which are critical for accuracy are preserved. 

The need for loss scaling can be avoided by using BFLOAT16 datatype.
The hardware numerics of BFLOAT16 on Intel architecture is available at~\cite{bfloat16}.
BFLOAT16 has been underlined to be a crucial ingredient for achieving peta-FLOPS scale on image classification task in~\cite{yingImage}.
In~\cite{cloudTPU}, Google notes the speed ups achieved on using this datatype over FP32 for TensorFlow models for various tasks, such as, image classification, image segmentation, object detection, machine translation.
It may be noted that the benefits of BFLOAT16 are not only restricted to machine learning paradigm, the recently published work~\cite{mcBfloat16} uses this representation for performing Monte Carlo simulations of Ising model which is an important model is statistical physics.
In fact, the authors of~\cite{mcBfloat16} mention BFLOAT16 ``provides better training and model accuracy than the IEEE half-precision representation''.
The Julia language~\cite{julia}, which is designed to provide high-performance without sacrificing ease of programming, has also been shown to be benefited from BFLOAT16~\cite{juliaBfloat16}.
OpenAI has also mentioned that optimizing kernels targeting BFLOAT16 is ``in active development''~\cite{openaiBfloat16}.


\section{Training with Brain Floating Point} 
\label{numerics}
Numerous studies have shown that 16-bits of precision is sufficient for training deep neural networks\cite{gupta2015lp},\cite{nvbaidu_mixed},\cite{flexpoint},\cite{mixedPrec18}. Researchers have experimented with various numeric formats to optimize training platforms for power and performance.  State-of-the-art training platforms today have chosen IEEE-754 half-precision floating point as the preferred numeric format for deep leaning training. However, the narrow dynamic range of half-precision floating point is not sufficient to represent error gradients during back propagation. To mitigate this, training methods use loss scaling techniques\cite{nvbaidu_mixed} to shift gradients into expressible range supported by half-precision floats. While this is easier to implement for certain feed-forward networks as a simple multiplication of loss with a constant scaling factor, others (e.g. recurrent) require a more sophisticated approach to determine the right scaling parameter. This process is often iterative and requires significant time and resource investment from data scientists to optimize it to their network. These software overheads become a hindrance for seamless migration of new deep learning applications to take advantage of the low-precision hardware. Recent developments in software tools such as ``automatic mixed precision''~\cite{nvidia_amp} are aimed at easing some of this burden from data scientists. However, these tools in their current form also require changes in the original model code and are not guaranteed to result in sufficient performance gains. 

The values are represented as truncated full precision floating point values with 8 bits of mantissa and the dynamic range comparable to FP32 (Table~\ref{tab:float_comparison}). The extended dynamic range can now represent smaller gradient values without applying complicated loss scaling methods, which enables easier migration of deep learning workloads to BFLOAT16 hardware. There are some additional benefits to adopting BFLOAT16 numeric format to build hardware for deep learning. Core compute primitives such as FMA can be built using 8-bit multipliers which lead to significant area and power savings while preserving the full dynamic range of FP32. 
Table~\ref{tab:float_comparison} shows the comparison of BFLOAT16 with other standard IEEE floating point formats. 

\begin{table}[!htbp] 
  \caption{Comparison BFLOAT16 numeric format with IEEE-754 FP32 and FP16 formats.}
  \label{tab:float_comparison}
  \centering
  \begin{tabular}{llllll}
    \toprule
    Data Type   & Bit Format    & Max    & Min      & Min        & Acc.  \\
                & (s, e, m)     & Normal & Normal   & Subnormal  &  Size\\
    \midrule 
    FP32     & 1, 8, 23  & \num{3.40e+38} & \num{1.17e-38}  & \num{1.40e-45}  & float32\\
    FP16     & 1, 5, 10  & \num{6.55e4}  & \num{6.10e-5}  & \num{5.96e-8}  & float32\\
    BFLOAT16             & 1, 8, 7   & \num{3.38e38}  & \num{1.17e-38}  & N/A   & float32\\
    \bottomrule
  \end{tabular}
\end{table}


\begin{figure}[h]
\centering
\includegraphics[width=0.8\textwidth,keepaspectratio]{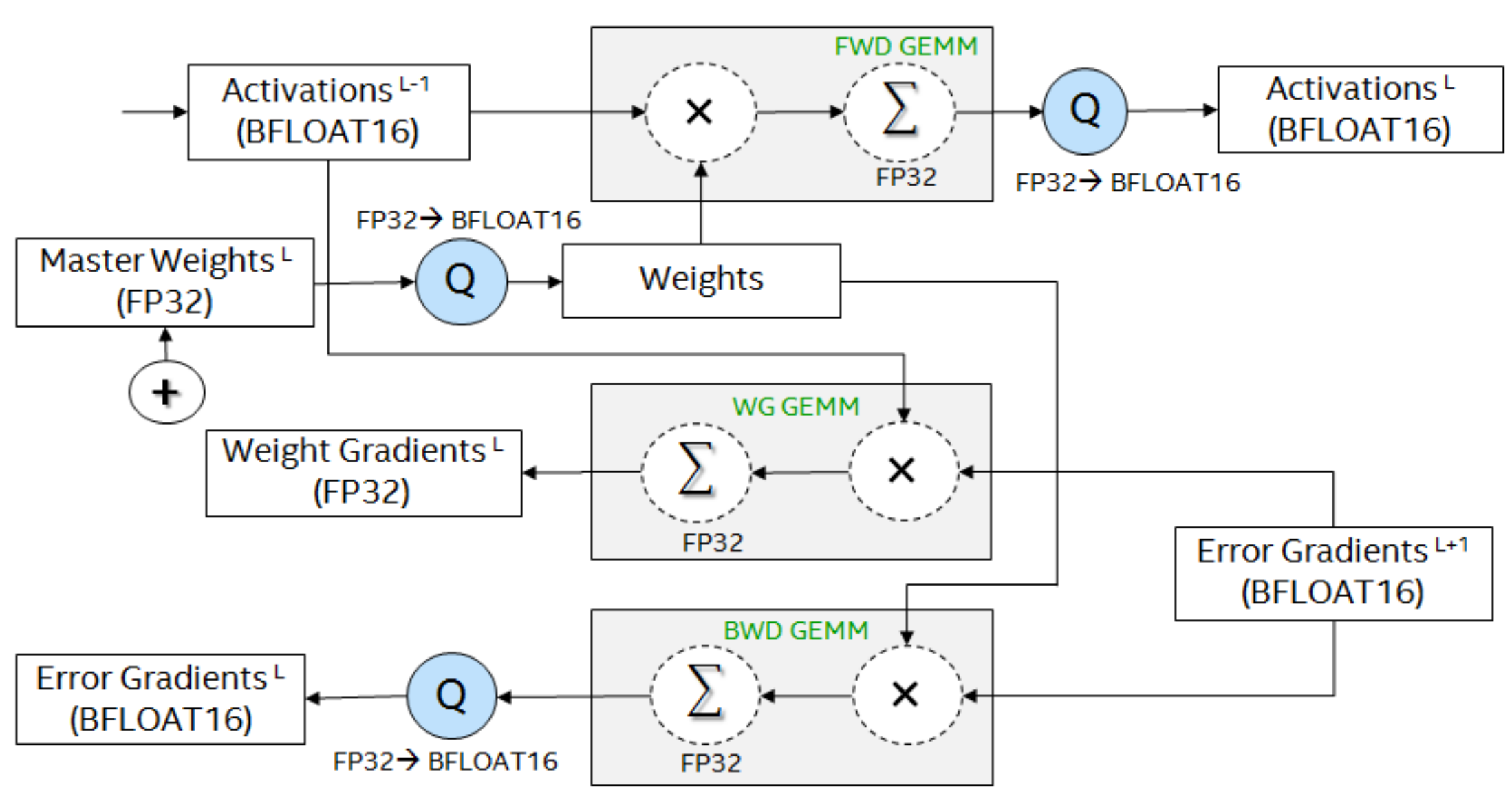}
\caption{Mixed precision data flow used for training DNNs with BFLOAT16 data format}
\label{fig:data_flow}
\end{figure}

Figure~\ref{fig:data_flow} shows the mixed precision data flow used to train deep neural networks using BFLOAT16 numeric format. The core compute kernels represented as GEMM operations accept inputs as BFLOAT16 tensors and accumulate the output to FP32 tensors. Quantlib (shown as Q in Figure~\ref{fig:data_flow}) modifies these output tensors to BFLOAT16 format before passing them to the next layer. Quantlib is also employed to modify a copy of the FP32 weights to BFLOAT16 for the forward pass. Error gradients with respect to the inputs also in BFLOAT16 format.
Non-GEMM compute operations including batch-normalization, and activation functions such as ReLU, tanh and sigmoid also accept BFLOAT16 tensors as inputs. Bias tensors are always maintained in FP32. The weight update step (e.g., in SGD solver) uses the FP32 copy of the weights to maintain model accuracy.   


\section {Results}
\label{results}
Our evaluation of BFLOAT16 consists of the aforementioned deep learning models from different application domains and frameworks, using the tensor modification method via Quantlib discussed in section~\ref{intro}.


\subsection{Convolution Neural Networks}
Convolutional neural networks (CNN) have been primarily used for computer vision applications such as image classification, object detection and semantic segmentation. CNNs have been extensively studied both in academia and industry, primarily driven by public benchmarks such as the ImageNet Large Scale Visual Recognition Competition (ILSVRC). Over the past few years the CNNs which have won the ILSVRC competition, have become well established benchmarks. Here we choose AlexNet (ILSVRC 2012)~\cite{alexnet} and ResNet-50 (ILSVRC 2015)~\cite{resnet}
 as representative models for the BFLOAT16 evaluation. 
In addition to Convolution and InnerProduct layers 
(which contributes to majority of the computations), we use BFLOAT16 emulations for ReLU, BatchNorm, Pooling, Dropout and EltWise 
layers as well. This ensures that the full training pipeline uses BFLOAT16, not necessitating the use of higher precision for the intermediate tensor outputs. 
\subsubsection{AlexNet}

\begin{figure}[htp]
\centering
\begin{subfigure}{.48\textwidth}
  \centering
  \setlength{\fboxsep}{2pt}\fbox{\includegraphics[width=0.95\linewidth,keepaspectratio]{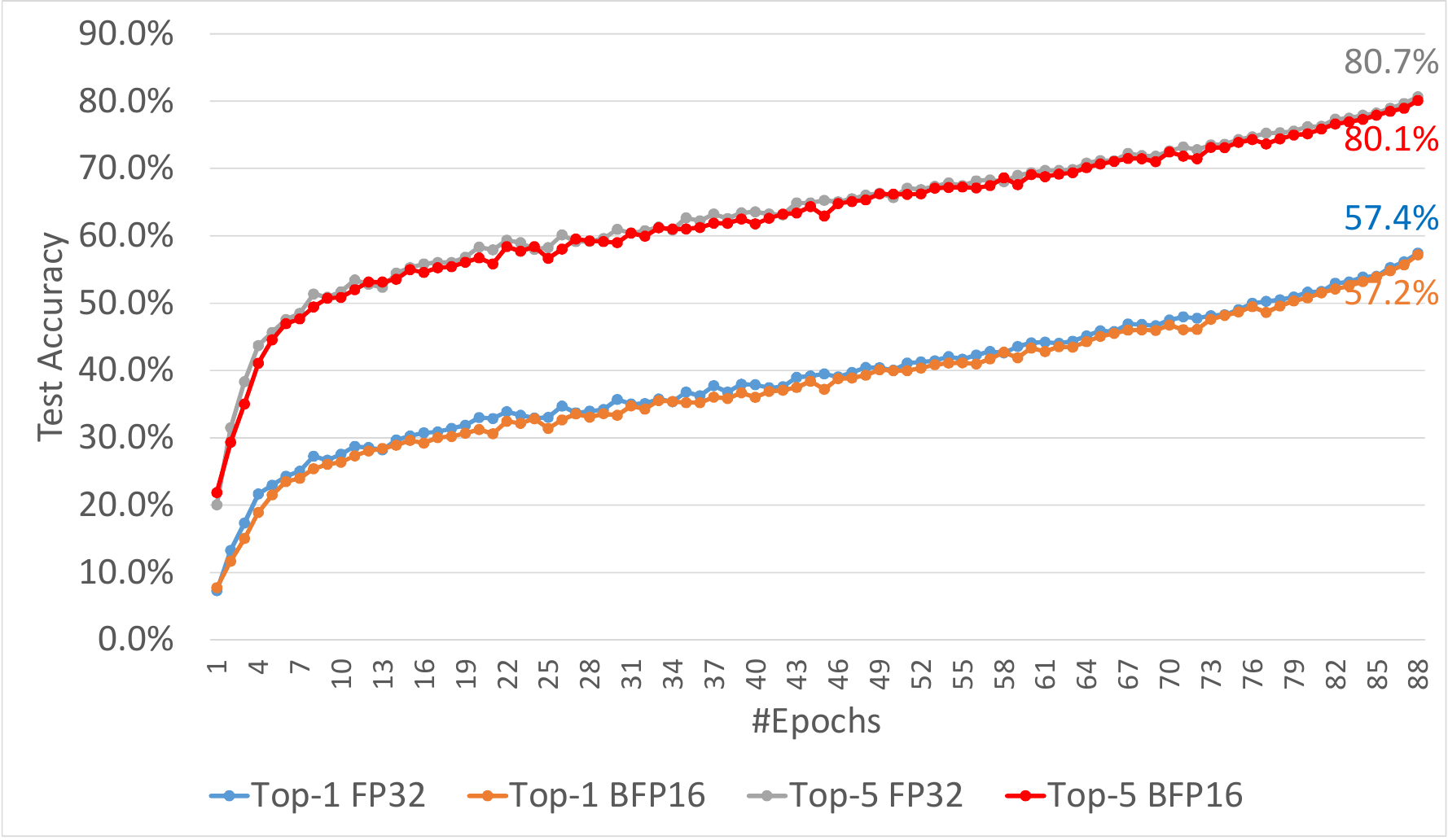}}
 \caption{AlexNet}
  \label{fig:alexnet_plot}
\end{subfigure}
\begin{subfigure}{.48\textwidth}
  \centering
  \setlength{\fboxsep}{2pt}\fbox{\includegraphics[width=0.95\linewidth,keepaspectratio]{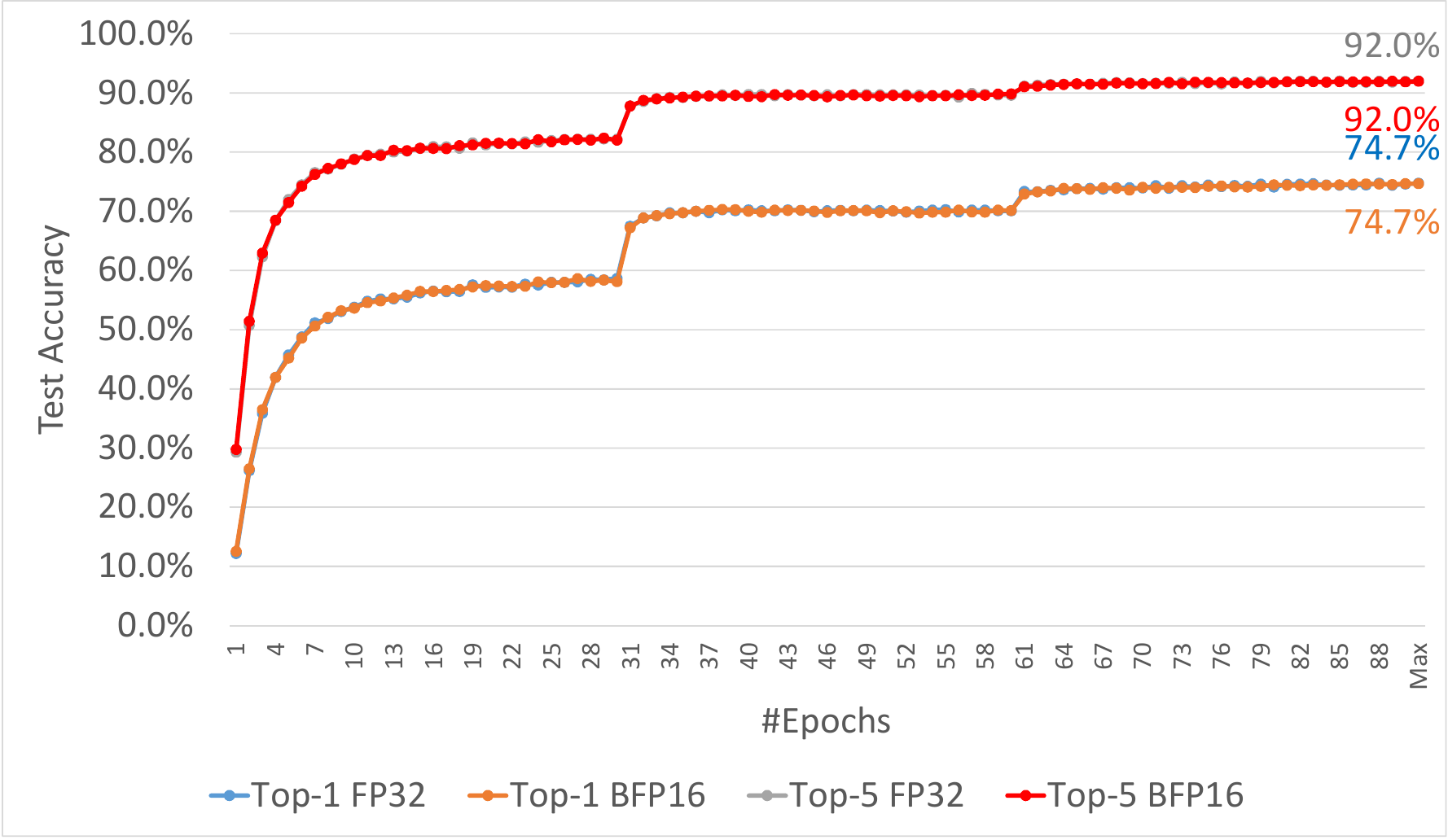}}
  \caption{ResNet-50}
  \label{fig:resnet_plot}
\end{subfigure}
\caption{Imagenet-1K training, top-1 and top-5 validation accuracy plots for CNNs
}
\label{fig:alexnet_resnet}
\end{figure}

For AlexNet, we used a global minibatch of 1024 running data parallel on 16 nodes for 88 epochs and achieved 57.4\% top-1 and 80.7\% top-5 accuracy. As shown in Figure~\ref{fig:alexnet_resnet}, our BFLOAT16 emulation follows very closely to the actual FP32 run and achieves 57.2\% top-1 and 80.1\% top-5 accuracy.


\subsubsection{ResNet}
Our Resnet-50 experiments are with a global minibatch of 1024 running data parallel on 32 nodes using SGD with Nestrov momentum. We trained for 90 epochs with learning rate warm up for first 5 epochs. Baseline FP32 run achieved top-1 accuracy of 74.7\% and top-5 accuracy of 92.0\% and as shown in Figure~\ref{fig:alexnet_resnet}, our BFLOAT16 emulation follows the baseline almost exactly and achieving the same top-1 and top-5 accuracy. During training we use local batch statistics for the batch normalization to compute validation accuracy after every epoch. The fully trained BFLOAT16 model, achieves 75.7\% top-1 test accuracy with global sample statistics, matching the baseline FP32 results.


\subsection{Recurrent Neural Networks}
Recurrent neural networks (RNN) unlike the feedforward networks allows for capturing temporal information due to its feedback connections. These models have been popularly used for applications such as automatic speech recognition (ASR) and language processing, which primarily involve sequence-based learning. RNNs have been observed to have more demanding numerical range requirements \cite{nvbaidu_mixed} and are more sensitive to the half precision datatype. For this class of networks we identify Baidu's DeepSpeech2~\cite{amodei2016deep} and Google's neural machine translation (GNMT) model~\cite{wu2016google} as representative candidates for the BFLOAT16 evaluation.

\subsubsection{DeepSpeech2}

The Deep speech 2 (DS2) topology,  consists of two convolution layers followed by 3 bi-directional gated recurrent unit (GRU) layers with 2048 cells and a final inner-product layer as a classifier. We use Adam optimizer to compute connectionist temporal classification loss (CTC)~\cite{ctc}. We use a batch size of 64 and a learning rate of 0.0005. The aforementioned model is trained on the librispeech dataset \cite{librispeech}, which consists of 460 hour corpus. 

\begin{figure}[htbp]
\centering
\begin{subfigure}{.47\textwidth}
  \centering
  \setlength{\fboxsep}{0pt}\fbox{\includegraphics[width=0.95\linewidth,keepaspectratio]{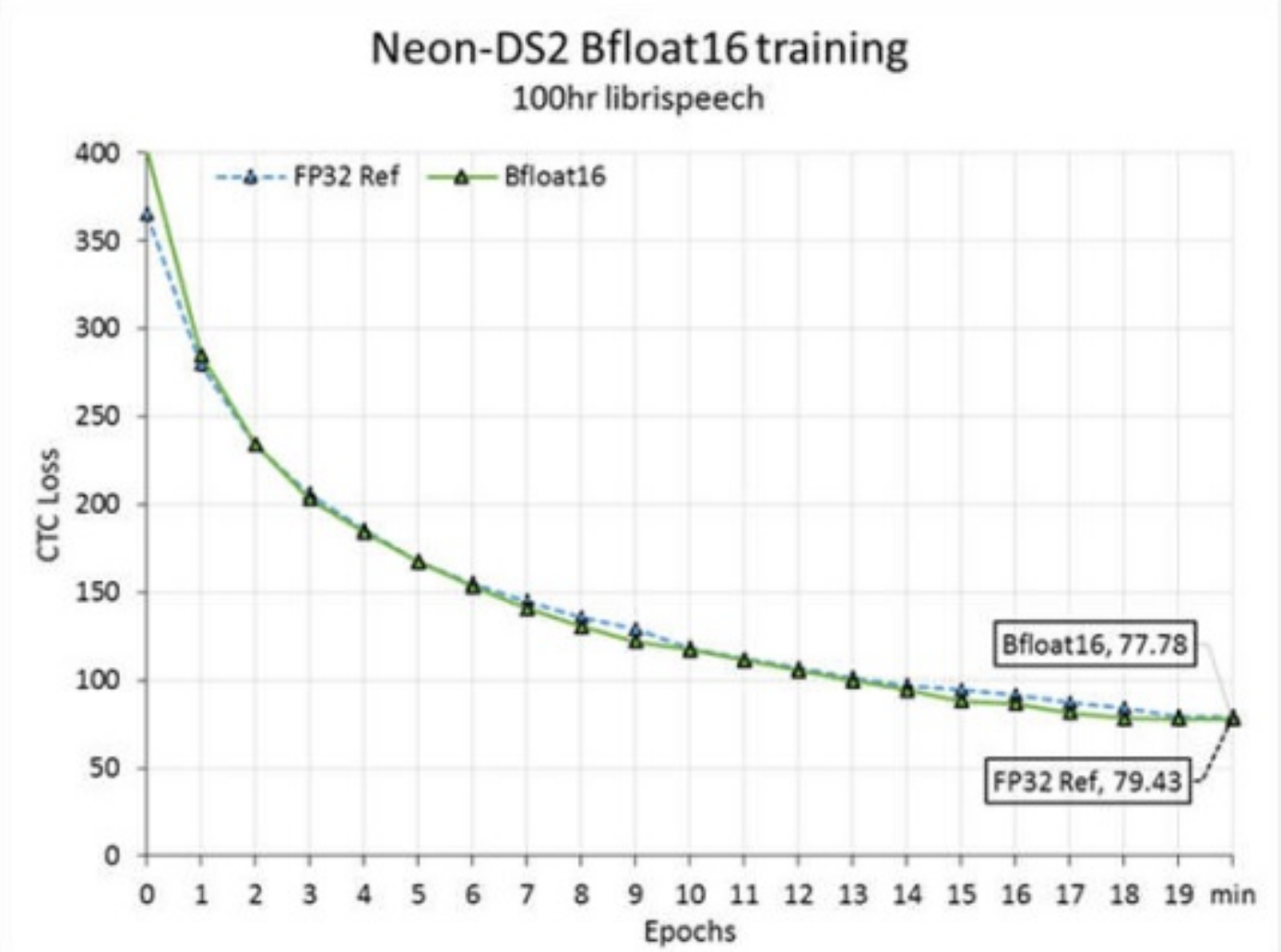}}
 \caption{DeepSpeech2}
  \label{fig:ds2_plot}
\end{subfigure}
\begin{subfigure}{.52\textwidth}
  \centering
  \setlength{\fboxsep}{6pt}\fbox{\includegraphics[width=0.94\linewidth,keepaspectratio]{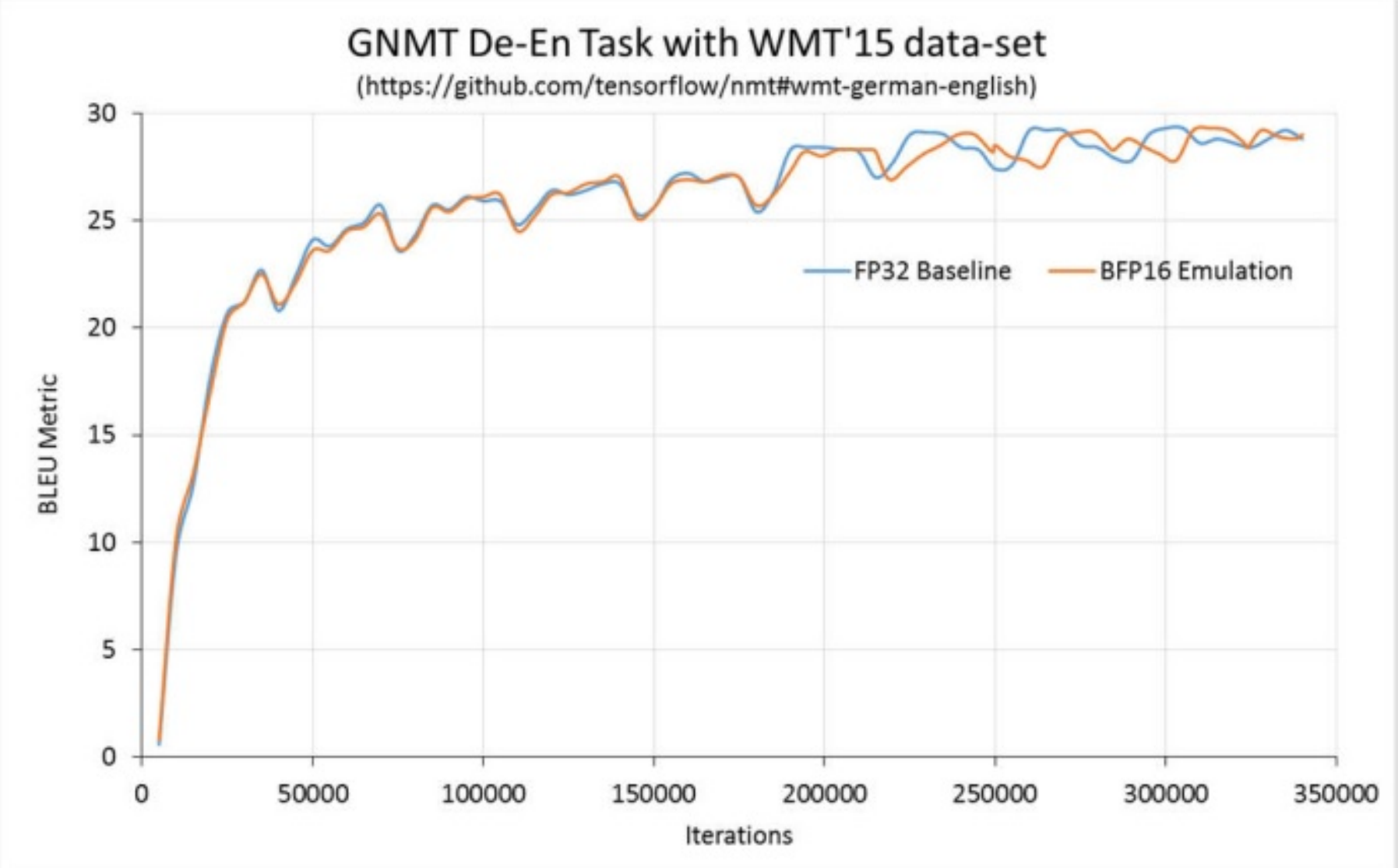}}
  \caption{GNMT}
  \label{fig:gnmt_plot}
\end{subfigure}
\caption{RNN training using BFLOAT16 data type.}
\label{fig:ds2_gnmt}
\end{figure}


\subsubsection{Neural Machine Translation}
Google's Neural Machine Translation (GNMT) is the SOTA neural machine translation model using a recurrent network. It uses stack of long short-term memory (LSTM) layers, along with an attention model for language modeling and translation. Table~\ref{gnmt_result_table} compares translation accuracy in terms of achieved BLEU scores for baseline FP32 and BFLOAT16 emulation. We use the small Vietnamese (VI) to English (EN) model and big German (DE) to English (EN) model. BFLOAT16 emulation achieves same or better accuracy than baseline. Figure~\ref{fig:ds2_gnmt} shows how closely BFLOAT16 emulation run follows the baseline FP32 run.

\begin{table}[h]
\caption{GNMT BLEU scores for De$\rightarrow$En and Vi$\rightarrow$En on WMT'16 and IWSLT'15 datasets.}
\label{gnmt_result_table}
\begin{center}
\begin{small}
\begin{sc}
\begin{tabular}{lcccr}
\toprule
Task                                        & FP32 & BFLOAT16  \\
\midrule
De$\rightarrow$En, WMT'16                   & 29.3 & 29.3 \\
Vi$\rightarrow$En, IWSLT'15 $+$ Attention   &17.1  & 18.3 \\
\bottomrule
\end{tabular}
\end{sc}
\end{small}
\end{center}
\vskip -0.1in
\end{table}

\subsection{Generative Adversarial Networks (GANs)}
Generative Adversarial Networks (GANs) have become a very important class of networks, they can be used to learn and mimic any arbitrary distribution of data. GANs achieve this by using two separate generator and discriminators networks in a tightly coupled way. Because GANs combine regression and discrimination tasks during training they tend to have different requirements for numerical precision and range. For the BFLOAT16 evaluation we consider DC-GAN\cite{dcgan} and SR-GAN \cite{ledig2017photo} models.
\subsubsection{DC-GAN}
DC-GAN~\cite{dcgan} represents a critical step in designing GAN architectures which were earlier known to be notoriously difficult to train.
This consists of fractionally-strided convolutions with ReLU activations in the generator, whereas convolutions with leaky ReLU
activations are used in the discriminator; batch normalization layers are used in both the generator and the discriminator.
We have implemented DC-GAN in Caffe and for our experiments all the input tensors (activations, weights) are converted to BFLOAT16 for
convolution layers (in both the generator and the discriminator), while only the input activations are converted to BFLOAT16 for batch normalization
layers; all other tensors are maintained in full precision.

A comparison between FP32 and BFLOAT16 is shown in Table~\ref{Tab:dcgan} in terms of inception scores and MS-SSIM.
As evident from the table, the outputs obtained for FP32 and BFLOAT16 are comparable.


\begin{table}[tbh]
 \begin{center}
 \caption{\label{Tab:dcgan}Comparison between FP32 and BFLOAT16 for DC-GAN on face dataset}
 \begin{tabular}{l r r}
\toprule
Datatype        & Inception Score & MS-SSIM \\
\midrule
Baseline (FP32) & $1.97\pm0.054$  & 0.262   \\
BFLOAT16        & $2.06\pm0.055$  & 0.217   \\
\bottomrule
 \end{tabular}
 \end{center}
\end{table}

\subsubsection{SR-GAN}
SR-GAN generates photo-realistic high-resolution images by super-resolving from a single shot of the low-resolution image \cite{ledig2017photo}. The low-resolution images are scaled $4\times$ preserving the spatial features while minimizing the noise. The quality of the output is measured using SSIM (Structural Similarity) and MS-SSIM (multi-scale structural similarity) and PNSR (peak signal to noise ratio) metrics.  
The topology consists of a ``generator'' network based on Resnet architecture, and the ``discriminator'' consists of 8 convolution layers each followed by a batchnorm and LeakyReLu. The network uses a ``VGG loss'' function using a pre-trained VGG19 network~\cite{vgg}. 

For the BFLOAT16 experiments, we converted all the inputs tensors (weights, activations) at convolution layers to BFLOAT16, while keeping the rest of the layers (Batchnorm, ReLU, LeakyReLu and Eltwise) at full precision. 
\begin{table}[h]
\caption{SR-GAN model trained with BFLOAT16 on DIV2K (http://www.vision.ee.ethz.ch/ntire17/) dataset.}
\label{srgan_resut_table}
\begin{center}
\begin{small}
\begin{sc}
\begin{tabular}{lcccr}
\toprule

Datatype                 & PSNR & SSIM & MS-SSIM  \\
\midrule
Baseline (FP32)          & 26.1749  &	0.73753	& 0.99999 \\
BFLOAT16                 & 26.1415	&   0.74079	& 0.99999 \\
\bottomrule
\end{tabular}
\end{sc}
\end{small}
\end{center}
\vskip -0.1in
\end{table}

\begin{figure}[htp]
\centering
\begin{subfigure}{.497\textwidth}
  \centering
  \setlength{\fboxsep}{0pt}\fbox{\includegraphics[width=0.98\linewidth,keepaspectratio]{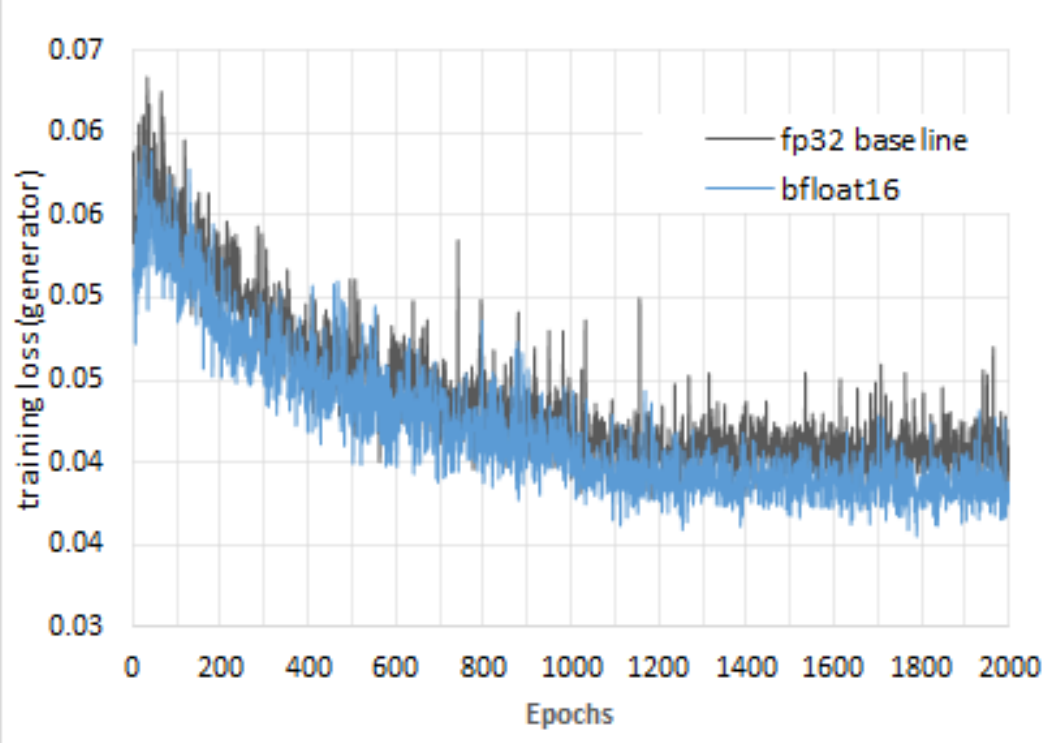}}
 \caption{Generator Network}
  \label{fig:srgan_g_loss}
\end{subfigure}
\begin{subfigure}{.497\textwidth}
  \centering
  \setlength{\fboxsep}{0pt}\fbox{\includegraphics[width=0.98\linewidth,keepaspectratio]{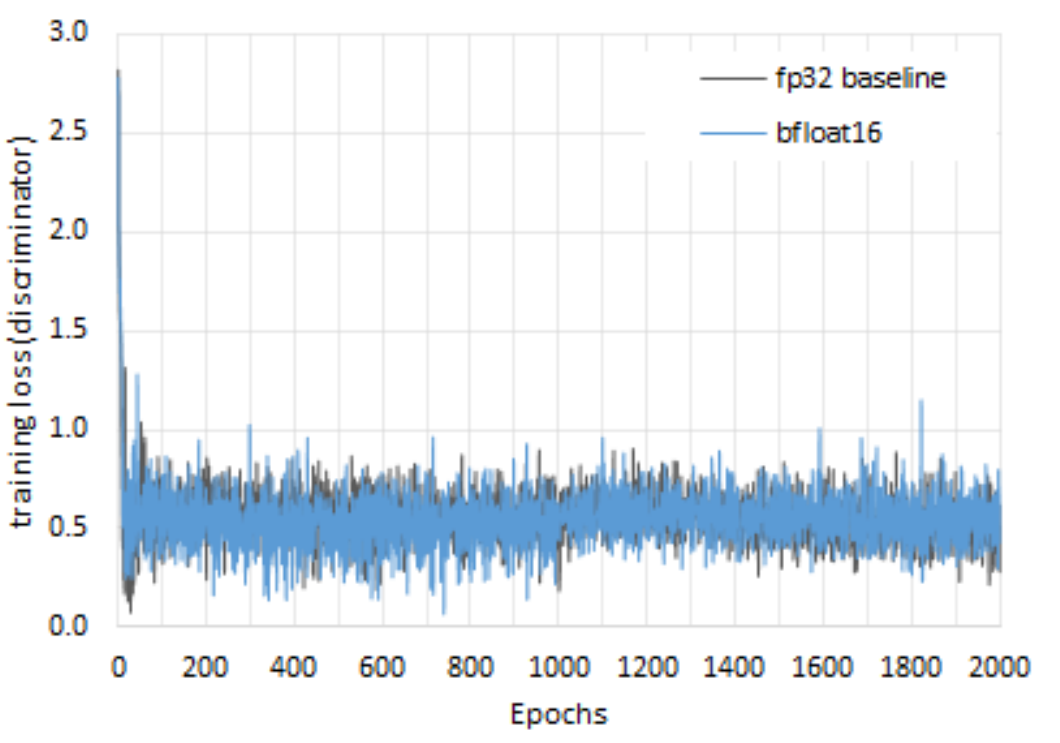}}
  \caption{Discriminator Network}
  \label{fig:srgan_d_loss}
\end{subfigure}
\caption{Training loss for SR-GAN training using BFLOAT16}
\label{fig:srgan_loss}
\end{figure}

\subsection{Industrial Scale Recommendation System}

Recommendation system and personalization models are very important for many practical-scale applications. Here we evaluate the Deep \& Cross Network~\cite{deepcross17} on a small Kaggle Criteo Dataset\footnote{https://www.kaggle.com/c/criteo-display-ad-challenge/data} and a typical DNN recommender system~\cite{fp16embeddings, dlrm} on a large Terabyte Criteo Dataset\footnote{https://www.criteo.com/news/press-releases/2015/07/criteo-releases-industrys-largest-ever-dataset/}, which target predicting the ads click-through rate~\cite{CTR}. 
The accuracy of the recommendation system models is measured by the log loss~\cite{deepcross17, fp16embeddings, dlrm}, which predicts when the users will click on ads. Lower log loss translates into higher prediction accuracy for the recommendation model. Note that an accuracy loss of 0.001 in log loss is considered unacceptable in practice.


For our BFLOAT16 experiments, all input tensors (activations and weights) are converted to BFLOAT16 for fully connected layers in both forward and backward propagation passes. During the weight update stages, we use a FP32 master copy~\cite{nvbaidu_mixed} to reduce the additional accuracy loss. We use either the round-to-nearest or direct truncation scheme when we do the conversion from BFLOAT16 to FP32.

The accuracy evaluation results are shown in Table \ref{recommender_resut_table}. As we can observe, BFLOAT16 with the round-to-nearest scheme is almost the same as FP32 baseline accuracy, while BFLOAT16 with the direct truncation scheme suffers from a tiny accuracy degradation ($\sim 0.02\%$).


\begin{table}[htbp]
\caption{The log loss for Deep \& Cross Network~\cite{deepcross17} on a small Kaggle Criteo dataset and a DNN recommender system~\cite{fp16embeddings} on a TeraByte Criteo dataset trained with BFLOAT16 (with either round-to-nearest or direct truncation for the conversion from FP32 to BFLOAT16).}
\label{recommender_resut_table}
\begin{center}
\begin{small}
\begin{sc}
\begin{tabular}{l @{\hskip 0.04in} c  c  cr}
\toprule
Recommendation system    & Baseline(FP32) & BFLOAT16 (rnd) & BFLOAT16 (trunc)  \\
\midrule
Deep \& Cross Network & 0.44372 & 0.44372 & 0.44393 \\
DNN Recommender System    & 0.12520  & 0.12520 & 0.12537 \\
\bottomrule
\end{tabular}
\end{sc}
\end{small}
\end{center}
\vskip -0.1in
\end{table}

\subsection{Beyond Emulation - Towards Bare Metal Execution}

We close this section by highlighting that our presented emulation strategy is an
excellent approximation of future Intel Xeon CPUs. We therefore took
the slim CNN training framework GxM, which was presented in~\cite{Georganas:2018:AHD:3291656.3291744}, and implemented all 
important operators (convolution, fully-connected, batch-normalization and pooling layers) 
utilizing the AVX512BF16 instruction set extensions~\cite{ise}. 
Therefore, all activation and weight
data was only present as 16bit data in memory and the special VNNI (vector neural network
instructions) datalayout was employed to support the BFLOAT16 dot-product instruction with
FP32 accumulation. The
execution of these instructions was done by bit-accurate emulation on current AVX512 silicon
with only a very slight performance tax. When training ResNet-50 on Imagenet
using the AVX512BF16 instructions, 
we achieved a Top-1 accuracy of 75.62\%, which matches the current state-of-the-art performance. 
Additionally, the code is already heavily optimized and
ready for prime-time. Additionally we have implemented a full BFLOAT16 LSTM-cell which is
currently being integrated into Tensorflow.

\section {Conclusion}
\label{conclusion}

Our goal in this paper was to establish BFLOAT16 as an alternative half-precision format for Deep Learning training, given that its dynamic range is the same as that of FP32 and conversion to/from FP32 is straightforward. Our empirical study demonstrates that, unlike IEEE 754 half-precision format and 16-bit Integer, BFLOAT16 based training eliminates the need for hyperparameter tuning or complex software management for block quantization. Our study also demonstrates that BFLOAT16 is a robust datatype having the ability to cover the range of tensors across application domains including vision, speech, language, generative networks, and recommendation systems. We expect industry-wide adoption of BFLOAT16 across emerging domains. 

\bibliographystyle{plain}
\bibliography{references}
\end{document}